\documentclass[11pt,a4paper]{article}
\usepackage{times,latexsym}
\usepackage{url}
\usepackage[T1]{fontenc}
\usepackage{algorithm}
\usepackage{algorithmic}
\usepackage{amsmath}
\usepackage{booktabs}
\usepackage{graphicx}
\usepackage{multirow}
\usepackage{makecell}
\usepackage[usenames,dvipsnames]{xcolor} 
\usepackage{wasysym}
\usepackage{paralist}
\usepackage[acceptedWithA]{tacl2021v1}

\usepackage{cleveref}

\usepackage{xspace,mfirstuc,tabulary}

\newif\iftaclinstructions
\taclinstructionsfalse % AUTHORS: do NOT set this to true
\iftaclinstructions
\renewcommand{\confidential}{}
\renewcommand{\anonsubtext}{(No author info supplied here, for consistency with
TACL-submission anonymization requirements)}
\newcommand{\instr}
\fi

\iftaclpubformat % this "if" is set by the choice of options

\else

\fi

\title{Meta-Learning a Cross-lingual Manifold for Semantic Parsing}

\author{Tom Sherborne \and Mirella Lapata\\
Institute for Language, Cognition and Computation\\
  School of Informatics, University of Edinburgh \\
10 Crichton  Street, Edinburgh EH8 9AB\\
  \texttt{tom.sherborne@ed.ac.uk,~mlap@inf.ed.ac.uk} \\
}

\date{}

\begin{document}
\maketitle
\begin{abstract}
  Localizing a semantic parser to support new languages requires
  effective cross-lingual generalization.  Recent work has found
  success with machine-translation or zero-shot methods although these
  approaches can struggle to model how native speakers ask questions.  
  We consider how to effectively leverage minimal annotated examples in new languages
  for few-shot cross-lingual semantic parsing. We introduce a first-order
  meta-learning algorithm to train a semantic parser with maximal
  sample efficiency during cross-lingual transfer.  Our algorithm uses
  high-resource languages to train the parser and simultaneously
  optimizes for cross-lingual generalization for lower-resource
  languages. Results across six languages on ATIS
  demonstrate that our combination of generalization steps yields
  accurate semantic parsers sampling~$\le$10\% of source training data in each new
  language. Our approach also trains a competitive 
  model on Spider using English with generalization to Chinese similarly sampling $\le$10\% of training data.\footnote{Our code
  and data are available at {\tt github.com/tomsherborne/xgr}}
\end{abstract}

% !TEX root = ../main.tex

\section{Introduction}
\label{sec:intro}

A semantic parser maps natural language (NL) utterances to logical
forms (LF) or executable programs in some machine-readable language
(e.g.,~SQL). Recent improvement in the capability of semantic parsers has
focused on domain transfer within English
\citep{su-yan-2017-xdomain-paraphrasing,suhr-etal-2020-exploring},
compositional generalization
\citep{yin-neubig-2017-syntactic-gcd-code-generation,
  herzig-berant-2021-span,scholak-etal-2021-picard}, and more recently
cross-lingual methods
\citep{multilingsp-and-code-switching-Duong2017,neural-hybrid-trees-susanto2017,polyglotapi-Richardson2018}.

Within cross-lingual semantic parsing, there has been an effort to
bootstrap parsers with minimal data to avoid the cost and labor
required to support new languages. Recent proposals include using
machine translation to approximate training data for supervised
learning
\citep{moradshahi-etal-2020-localizing,sherborne-etal-2020-bootstrapping,nicosia-etal-2021-translate-fill-zero-shot-semparse}
and zero-shot models, which engineer cross-lingual similarity with
auxiliary losses
\citep{van-der-goot-etal-2021-masked,yang-etal-2021-frustratingly-simple-DRS-parsing,sherborne-lapata-2022-zero}. These
shortcuts bypass costly data annotation but present limitations such
as ``translationese'' artifacts from machine translation
\citep{koppel-ordan-2011-translationese} or undesirable domain
shift \citep{sherborne-lapata-2022-zero}. However, annotating a minimally
sized data sample can potentially overcome these limitations while incurring
significantly reduced costs compared to full dataset translation \citep{garrette-baldridge-2013-learning}.

We argue that a few-shot approach is more realistic for an engineer
motivated to support additional languages for a database -- as one can
rapidly retrieve a high-quality sample of translations and
combine these with existing supported languages (i.e., English).
Beyond semantic parsing, cross-lingual few-shot approaches have also
succeeded at leveraging a small number of annotations within a variety
of tasks \citep[\emph{inter alia}]{zhao-etal-2021-closer} including
natural language inference, paraphrase identification,
part-of-speech-tagging, and named-entity recognition. Recently, the application
of meta-learning to domain generalization has further demonstrated capability
for models to adapt to new domains with small samples
\citep{gu-etal-2018-meta-nmt,li-hospedales-metadg,wang-etal-fewshot-survey}.

In this work, we synthesize these directions into a meta-learning
algorithm for cross-lingual semantic parsing. Our approach 
explicitly optimizes cross-lingual generalization using fewer training samples per new language without performance degradation. We also require minimal computational overhead beyond standard gradient-descent training and no external dependencies beyond
in-task data and a pre-trained encoder.  Our algorithm,
\textbf{Cross}-Lingual \textbf{G}eneralization \textbf{Reptile}
(\mbox{\sc XG-Reptile}) unifies two-stage meta-learning into a single
process and outperforms prior and constituent methods on all
languages, given identical data constraints. The
proposed algorithm is still model-agnostic and applicable to more tasks 
requiring sample-efficient cross-lingual transfer.

Our innovation is the combination of both \mbox{intra-task} and
\mbox{inter-language} steps to jointly learn the parsing task and optimal
cross-lingual transfer. Specifically, we interleave learning 
the overall task from a high-resource language and learning cross-lingual
transfer from a minimal sample of a lower-resource language. Results
on ATIS \citep{hemphill-etal-1990-atis} in six languages (English,
French, Portuguese, Spanish, German, Chinese) and Spider
\citep{yu-etal-2018-spider} in two languages (English, Chinese)
demonstrate our proposal works in both single- and cross-domain
environments. Our contributions are as follows: 

\vspace{-0.5em}
\begin{itemize}
\item We introduce \mbox{\sc XG-Reptile}, a
    first-order meta-learning algorithm for cross-lingual generalization.
    \mbox{\sc XG-Reptile} approximates an \textit{optimal manifold} using support
    languages with \textit{cross-lingual regularization} using target
    languages to train for explicit cross-lingual similarity.
    
\item We showcase sample-efficient cross-lingual transfer within 
    two challenging semantic parsing datasets across multiple languages.
    Our approach yields more accurate parsing in a few-shot scenario
    and demands 10$\times$ fewer samples than prior methods. 
    
\item We establish a cross-domain and cross-lingual 
    parser obtaining promising results for both Spider in English 
    \citep{yu-etal-2018-spider} and CSpider in Chinese \citep{Min2019-CSPIDER}.
\end{itemize}

%%% Local Variables: 
%%% mode: latex
%%% TeX-master:  "../main"
%%% End:

% !TEX root = ../main.tex

\section{Related Work}
\label{sec:rw}

\paragraph{Meta-Learning for Generalization}

Meta-Learning\footnote{We refer the interested reader to
\citet{wang-etal-fewshot-survey},
\citet{hospedales-meta-learning-survey}, and
\citet{Wang2021GeneralizingTU-survey} for more extensive surveys on
meta-learning.}  has recently emerged as a promising technique for
generalization, delivering high performance on unseen domains by
\emph{learning to learn}, i.e., improving learning over multiple
episodes
\citep{hospedales-meta-learning-survey,Wang2021GeneralizingTU-survey}. A popular approach is Model-Agnostic Meta-Learning
\citep[MAML]{pmlr-v70-finn17a}, wherein the goal is to train a model
on a variety of learning tasks, such that it can solve \emph{new}
tasks using a small number of training samples. In effect, MAML
facilitates task-specific fine-tuning using few samples in
a two-stage process. MAML requires computing higher-order gradients
(i.e.,~``gradient through a gradient'') which can often be
prohibitively expensive for complex models. This limitation has motivated
\textit{first-order} approaches to MAML which offer similar
performance with improved computational efficiency.

In this vein, the {Reptile} algorithm
\citep{DBLP:journals/corr/abs-1803-02999-REPTILE} transforms the
higher-order gradient approach into $K$ successive
first-order steps. 
{Reptile}-based training learns an \textit{optimal manifold} across tasks (i.e., a high-density parameter sub-region biased for strong cross-task likelihood), then
similarly followed by rapid fine-tuning. By learning an optimal 
initialization, meta-learning proves useful for low-resource
adaptation by minimizing the data required for out-of-domain tuning on new tasks. \citet{kedia-etal-2021-beyond-reptile} also demonstrate the utility of Reptile to improve \emph{single-task} performance. We build on this to examine single-task cross-lingual transfer using the \textit{optimal manifold} learned with Reptile.

\paragraph{Meta-Learning for Semantic Parsing}
A variety of NLP applications have adopted meta-learning in zero- and
few-shot learning scenarios as a method of explicitly training for
generalization \citep{lee-etal-2021-meta-tutorial,
  hedderich-etal-2021-survey-metanlp}. Within semantic parsing, there
has been increasing interest in \textit{cross-database generalization}, motivated by datasets such as Spider
\citep{yu-etal-2018-spider} requiring navigation of unseen
databases
\citep{sp-over-many-kbs-Herzig2017, suhr-etal-2020-exploring}.

Approaches to generalization have included simulating source and
target domains \citep{givoli-reichart-2019-zero-shot-semantic-parsing}
and synthesizing new training data based on unseen databases
\citep{zhong-etal-2020-grounded-adaptation-gazp,
  xu-etal-2020-autoqa}. Meta-learning has demonstrated fast adaptation
to new data within a monolingual low-resource setting
\citep{huang-etal-2018-natural-structured-query-meta-learning-parsing,guo-etal-2019-coupling-context-meta-learning-parsing,DBLP:journals/corr/abs-1905-11499,Sun_Tang_Duan_Gong_Feng_Qin_Jiang_2020-meta-learning-parsing}. Similarly,
\citet{chen-etal-2020-low-resource-domain-adaptation-task-oriented-parsing}
utilize Reptile to improve generalization of a model, trained
on source domains, to fine-tune on new domains. Our
work builds on \citet{wang-etal-2021-meta-DGMAML}, who explicitly
promote monolingual cross-domain generalization by ``meta-generalizing'' across
disjoint domain-specific batches during training.

\paragraph{Cross-lingual Semantic Parsing}

A surge of interest in cross-lingual NLU has seen the creation of many
benchmarks across a breadth of languages
\citep{conneau2018xnli,hu2020xtreme,liang-etal-2020-xglue}, thereby
motivating significant exploration of cross-lingual transfer
\citep[\emph{inter alia}]{nooralahzadeh-etal-2020-zero-xmaml,xia-etal-2021-metaxl,
  xu-etal-2021-soft-layer-selection-meta-simpler-xmaml,zhao-etal-2021-closer}.
Previous approaches to cross-lingual semantic parsing assume
parallel multilingual training data
\citep{multilingual-sp-hierch-tree-Jie2014} and exploit multi-language
inputs for training without resource constraints
\citep{arch-for-neural-multisp-Susanto2017,
  neural-hybrid-trees-susanto2017}.

There has been recent interest in evaluating if machine translation is
an economic proxy for creating training data in new languages
\citep{sherborne-etal-2020-bootstrapping,moradshahi-etal-2020-localizing}.
Zero-shot approaches to cross-lingual parsing have also been explored using
auxiliary training objectives
\cite{yang-etal-2021-frustratingly-simple-DRS-parsing,sherborne-lapata-2022-zero}.
Cross-lingual learning has also been gaining traction in the adjacent
field of spoken-language understanding (SLU).
For datasets such as MultiATIS \citep{Upadhyay2018-multiatis},
MultiATIS++ \citep{xu-etal-2020-end-multiatis}, and MTOP
\citep{li-etal-2021-mtop}, zero-shot cross-lingual transfer has been
studied through specialized decoding methods
\citep{zhu-etal-2020-dont-parse-insert}, machine translation
\citep{nicosia-etal-2021-translate-fill-zero-shot-semparse}, and
auxiliary objectives \citep{van-der-goot-etal-2021-masked}.

Cross-lingual semantic parsing has mostly remained orthogonal to the
cross-database generalization challenges raised by datasets such as
Spider \citep{yu-etal-2018-spider}. While we primarily present
findings for multilingual ATIS into SQL 
\citep{hemphill-etal-1990-atis}, we also train a parser on both Spider
and its Chinese version \citep{Min2019-CSPIDER}. To the best of our
knowledge, we are the first to explore a multilingual approach to this
cross-database benchmark. We use {Reptile} to learn the overall task
and leverage domain generalization techniques
\citep{li-hospedales-metadg, wang-etal-2021-meta-DGMAML} for
sample-efficient cross-lingual transfer.

%%% Local Variables: 
%%% mode: latex
%%% TeX-master:  "../main"
%%% End:

% !TEX root = ../main.tex

\section{Problem Definition}
\label{sec:problem}

\paragraph{Semantic Parsing}

We wish to learn a parameterized parsing function, $p_{\theta}$, which
maps from a natural language utterance and a relational database
context to an executable program expressed in a logical form (LF) language:
\begin{align}
P = p_{\theta}\left(Q,~D\right) \label{eq:parser_fn}
\end{align}

As formalized in Equation~\eqref{eq:parser_fn}, we learn
parameters,~$\theta$, using paired data
$\left(Q,~P,~\mathcal{D}\right)$ where~$P$ is the logical form
equivalent of natural language question~$Q$. In this work, our LFs are
all executable SQL queries and therefore grounded in a
database~$\mathcal{D}$. A single-domain dataset references only
one~$\mathcal{D}$ database for all $\left(Q,~P\right)$, whereas a
multi-domain dataset demands reasoning about unseen databases to
generalize to new queries. This is expressed as a `zero-shot'
problem if the databases at test time, $\mathcal{D}_{\rm test}$, were
unseen during training. This challenge demands a parser
capable of \textit{domain generalization} beyond observed
databases. This is in addition to the \textit{structured prediction}
challenge of semantic parsing.

\paragraph{Cross-Lingual Generalization}

Prototypical semantic parsing datasets express the question, ${Q}$, in
English only. As discussed in Section~\ref{sec:intro}, our parser
should be capable of mapping from \emph{additional} languages to
well-formed, executable programs. However, prohibitive expense limits
us from reproducing a monolingual model for each additional language
and previous work demonstrates accuracy improvement by training
\textit{multilingual} models
\citep{multilingual-sp-hierch-tree-Jie2014}. In addition to the
challenges of structured prediction and domain generalization, we
jointly consider \textit{cross-lingual generalization}. Training
primarily relies on existing English data (i.e., ${Q}_{\rm EN}$
samples) and we show that our meta-learning algorithm in
Section~\ref{sec:methodology} leverages a small sample of training data in new
languages for accurate parsing. We  express this sample, $\mathcal{S}_{l}$, for some language, $l$,
as: 
\begin{align}
	\mathcal{S}_{l}&=\left({Q}_{l},~{P},~\mathcal{D}\right)_{i=0}^{N_{l}} \label{eq:sample_l}
\end{align} 
where $N_{l}$ is the sample size from~$l$, assumed to be smaller than the
original English dataset (i.e.,~\mbox{${N}_{l} \ll N_{\rm
    EN}$}). Where available, we extend this paradigm to develop models
for $L$ different languages simultaneously in a multilingual setup by
combining samples as:
\begin{align}
	\mathcal{S}_{\rm L} &= \lbrace\mathcal{S}_{ l_1},\mathcal{S}_{ l_2},\ldots,\mathcal{S}_{l_N}\rbrace \label{eq:multilingual_s_outer}
\end{align}
We can express cross-lingual generalization as:
\begin{align}
	 p_{\theta}\left(P~|~Q_{l},~\mathcal{D}\right) \rightarrow p_{\theta}\left(P~|~Q_{\rm EN},~\mathcal{D}\right) \label{eq:xlingual_distributions_approach}
\end{align}
where~$p_{\theta}\left(P~|~Q_{\rm EN},~\mathcal{D}\right)$ is the
predicted distribution over all possible output SQL sequences
conditioned on an English question, $Q_{\rm EN}$, and a
database~$\mathcal{D}$. Our goal is for the prediction from a new language, $Q_{l}$, to converge
towards this existing distribution using the same parameters $\theta$, constrained to fewer
samples in~$l$ than English.

We aim to maximize the accuracy of predicting programs on
unseen test data from each non-English language~$l$. The key challenge
is learning a performant distribution over each new language with
minimal available samples. This includes learning to incorporate
each $l$ into the parsing task and modeling the language-specific
surface form of questions. Our setup is akin to few-shot learning; however, the number of examples needed for satisfactory performance is
an empirical question. We are searching for both minimal sample
sizes and maximal sampling efficiency. We discuss our sampling
strategy in Section~\ref{sec:sampling_for_xgen} with results at
multiple sizes of~$\mathcal{S}_{L}$ in Section~\ref{sec:results}.

%%% Local Variables: 
%%% mode: latex
%%% TeX-master:  "../main"
%%% End:

% !TEX root = ../main.tex

\section{Methodology}
\label{sec:methodology}

% Background paragraph
We combine two meta-learning techniques for
cross-lingual semantic parsing. The first is the {Reptile} algorithm
outlined in Section~\ref{sec:rw}. {Reptile} optimizes for dense likelihood
regions within the parameters (i.e., an \emph{optimal manifold}) through
promoting inter-batch generalization \citep{DBLP:journals/corr/abs-1803-02999-REPTILE}.
Standard Reptile iteratively optimizes the manifold for an improved
initialization across objectives. Rapid fine-tuning yields the final task-specific model. The second technique is the first-order approximation of DG-MAML
\citep{li-hospedales-metadg,wang-etal-2021-meta-DGMAML}. This
single-stage process optimizes for domain generalization by simulating
``source'' and ``target'' batches from different domains to
explicitly optimize for \emph{cross-batch} generalization. Our
algorithm, \mbox{\sc XG-Reptile}, combines these paradigms to optimize
a target loss with the overall learning ``direction'' derived as
the \textit{optimal manifold} learned via {Reptile}. This trains 
an accurate parser demonstrating sample-efficient cross-lingual transfer
within an efficient \emph{single-stage} learning process.

\subsection{The \mbox{\sc XG-Reptile} Algorithm}

Each learning episode of \mbox{\sc XG-Reptile} comprises
two component steps: \mbox{\emph{intra-task}} learning and
\emph{inter-language} generalization to jointly learn parsing and
cross-lingual transfer. Alternating these processes trains a
competitive parser from multiple languages with low computational
overhead beyond existing gradient-descent training. Our approach
combines the typical two stages of meta-learning to produce a single
model without a fine-tuning requirement.

\begin{figure}[t]
    \centering
    \includegraphics[width=\columnwidth]{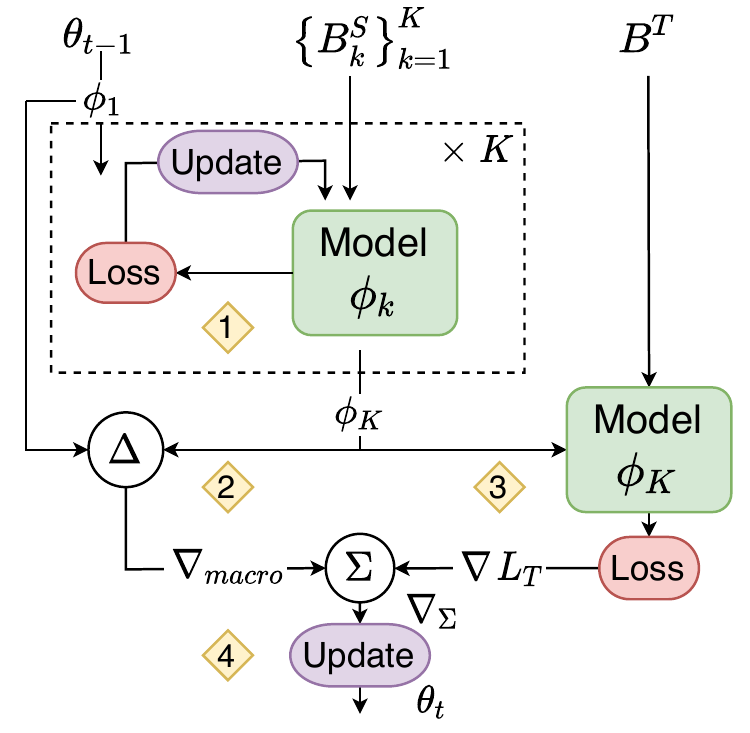}
    \vspace{-2.5em}
    \caption{\small One iteration of \mbox{\sc XG-Reptile}. (1) Run
      $K$ iterations of gradient descent over $K$ support batches to
      learn $\phi_{K}$, (2) compute $\nabla_{macro}$, the difference
      between $\phi_{K}$ and $\phi_{1}$, (3) find the loss on the
      target batch using $\phi_{K}$ and (4) compute the final gradient
      update from $\nabla_{macro}$ and the target loss.}
    \label{fig:xgr_diagram}
    \vspace{-1em}
\end{figure}

\paragraph{Task Learning Step}

We first sample from the high-resource language (i.e.,
$\mathcal{S}_{\rm EN}$) $K$ ``support'' batches of examples,
$\mathcal{B}^{S}=\lbrace\left(Q_{\rm
    EN},~P,~\mathcal{D}\right)\rbrace$. For each of $K$ batches: we
compute predictions, compute losses, calculate gradients and adjust
parameters using some optimizer (see illustration in
Figure~\ref{fig:xgr_diagram}). After $K$ successive optimization steps
the initial weights in this episode,~$\phi_{1}$, have been optimized
to~$\phi_{K}$. The difference between final and initial weights is
calculated as:

\vspace{-\baselineskip}
\begin{align}
    \nabla_{macro} = \phi_{K} - \phi_{1} \label{eq:reptile_step}
\end{align}

This ``macro-gradient'' step is equivalent to a {Reptile} step
\citep{DBLP:journals/corr/abs-1803-02999-REPTILE}, representing
learning an optimal \textit{manifold} as an approximation
of overall learning trajectory.

\paragraph{Cross-Lingual Step}

The second step samples one ``target'' batch,
$\mathcal{B}^{T}=\left(Q_{l},~P,~\mathcal{D}\right)$, from a
sampled target language (i.e., $\mathcal{S}_{
  l}\subset\mathcal{S}_{\rm L}$). We compute the cross-entropy
loss and gradients from the prediction of the model at $\phi_{K}$ on
$\mathcal{B}^{T}$:

\vspace{-\baselineskip}
\begin{align}
    \mathcal{L}_{T} = \textrm{Loss}\left(p_{\phi_{K}}\left(Q_{l},~\mathcal{D}\right),~P\right) \label{eq:cross_lingual_step}
\end{align}

We evaluate the parser at $\phi_{K}$ on a target
language we desire to generalize to. We show below that the
gradient of $\mathcal{L}_{T}$ comprises the loss at $\phi_{K}$ and
additional terms maximizing the inner product between the high-likelihood
manifold and the target loss. The total gradient encourages
intra-task and cross-lingual learning (see Figure~\ref{fig:xgr_diagram}).

\begin{algorithm}[!t]
\caption{{\sc XG-Reptile}\label{xgr-algo}}
{\small
\begin{algorithmic}[1]
    \REQUIRE{Support data, $\mathcal{S}_{\rm EN}$, target data, $\mathcal{S}_{L}$}
    \REQUIRE{Inner learning rate, $\alpha$, outer learning rate, $\beta$}
    \STATE{Initialise $\theta_{1}$, the vector of initial parameters}
    \FOR{$t \leftarrow 1$~\textbf{to}~$T$}
        \STATE{Copy $\phi_{1} \leftarrow \theta_{t-1}$}
        \STATE{Sample $K$ support batches $\lbrace\mathcal{B}^{S}\rbrace_{k=1}^{K}$ from $\mathcal{S}_{\rm EN}$}
        \STATE{Sample target language $l$ from $L$ languages}
        \STATE{Sample target batch~$\mathcal{B}^{T}$ from $\mathcal{S}_{l}$}
        \FOR{$k \leftarrow 1$~{\textbf{to}}~$K$ [Inner Loop]}
            \STATE{$\mathcal{L}^{S}_{k} \leftarrow {\rm Forward}\left(\mathcal{B}^{S}_{k},~\phi_{k-1}\right)$}
            \STATE{$\phi_{k}\leftarrow{\rm Adam}\left(\phi_{k-1},~\nabla\mathcal{L}^{S}_{k},~\alpha\right)$\label{algo:line_grad}} 
        \ENDFOR
        \STATE{Macro grad: ~$\nabla_{macro}\leftarrow\phi_{K}-\phi_{1}$}
        \STATE{Target Step: ~$\mathcal{L}_{T}\leftarrow {\rm Forward}\left(\mathcal{B}^{T},~\phi_{K}\right)$}
        \STATE{Total gradient: $\nabla_{\Sigma}=\nabla_{macro} + \nabla_{\phi_K}\mathcal{L}_{T}$}
        \STATE{Update $\theta_{t} \leftarrow {\rm SGD}\left(\theta_{t-1},~\nabla_{\Sigma},~\beta\right)$}
    \ENDFOR
\end{algorithmic}
}
\end{algorithm}

Algorithm \ref{xgr-algo} outlines the \mbox{\sc XG-Reptile} process (loss
calculation and batch processing are simplified for brevity). We repeat
this process over $T$ episodes to train model $p_{\theta}$ to
convergence. If we optimized for target data to align with individual
support batches (i.e., $K=1$) then we may observe batch-level noise in cross-lingual
generalization. Our intuition is that aligning the target gradient with an
approximation of the task manifold, i.e., $\nabla_{macro}$, will overcome this
noise and align new languages to a more mutually beneficial direction during
training. We observe this intuitive behavior during learning in Section~\ref{sec:results}.

We efficiently generalize to low-resource languages by
exploiting the asymmetric data requirements between steps: one
batch of the target language is required for $K$ batches of the source
language. For example, if $K=10$ then using this $\frac{1}{K}$
proportionality requires 10\% of target-language data relative to
support. We demonstrate in Section~\ref{sec:results} that we
can use a smaller $<\frac{1}{K}$ quantity per target language to 
increase sample efficiency. 
\vspace{-0.25em}

\paragraph{Gradient Analysis}

Following \citet{DBLP:journals/corr/abs-1803-02999-REPTILE}, we
express,~$g_{k}=\nabla\mathcal{L}^{S}_{k}$, the gradient in a single
step of the inner loop (Line~\ref{algo:line_grad}) as:

\vspace{-\baselineskip}
\begin{align}
\centering
g_{k} &= \bar{g}_i + \bar{H}_{k}\left( \phi_{k}-\phi_{1}\right) +
O\left(\alpha^2 \right) \label{eq:gi_gradient_taylor_2} 
\end{align}
We use a Taylor series expansion to approximate~$g_k$
by~$\bar{g}_{k}$, the gradient at the original point~$\phi_{1}$, the
Hessian matrix of the gradient at the initial point, $\bar{H}_{k}$,
the step difference between position~$\phi_{k}$ and the initial
position and some scalar terms with marginal influence,
$O\left(\alpha^{2}\right)$.

By evaluating Equation~\eqref{eq:gi_gradient_taylor_2} at $i=1$ and
rewriting the difference as a sum of gradient steps (e.g.,~Equations
\eqref{eq:initial_pt_trick} and \eqref{eq:diff_trick}), we arrive at
an expression for $g_{k}$ shown in Equation~\eqref{eq:gi_gradient_taylor_3} 
expressing the gradient as an initial component, $\hat{g}_k$,
and the product of the Hessian at~$k$, with all prior gradient steps. We
refer to \citet{DBLP:journals/corr/abs-1803-02999-REPTILE} for further
validation that the gradient of this product maximizes the
cross-batch expectation -- therefore promoting intra-task
generalization and learning the optimal manifold. The final gradient (Equation \eqref{eq:reptile_gradient_taylor})~
is the accumulation over $g_{k}$ steps and is equivalent to Equation \eqref{eq:reptile_step}. $\nabla_{macro}$ comprises both gradients of $K$ steps and
additional terms maximizing the inner-product of inter-batch gradients.

\vspace{-\baselineskip}
\begin{align}
{\rm Use}~~& g_{j}=\bar{g}_{j} + O\left(\alpha\right) \label{eq:initial_pt_trick} \\  
 &~\phi_{k}-\phi_{1}=-\alpha\sum_{j=1}^{k-1}g_{j} \label{eq:diff_trick}\\
g_{k} &= \bar{g}_i  - \alpha\bar{H}_{i}\sum_{j=1}^{k-1}\bar{g}_{j} + O\left(\alpha^2 \right) \label{eq:gi_gradient_taylor_3} \\
&\nabla_{macro} = \sum_{k=1}^{K}g_{k} \label{eq:reptile_gradient_taylor}  
\end{align}

We can similarly express the gradient of the target batch as Equation
\eqref{eq:target_gradient_taylor} where the term,
$\bar{H}_{T}\nabla_{macro}$, is the cross-lingual generalization
product similar to the intra-task generalization seen above.
\begin{align}
g_{T} &= \bar{g}_{T} -\alpha\bar{H}_{T}\nabla_{macro} + O\left(\alpha^2 \right) \label{eq:target_gradient_taylor} 
\end{align}

Equation~\eqref{eq:sigma_gradient_taylor} shows an example final \tabularnewline
gradient when $K=2$. Within the parentheses are
the intra-task and cross-lingual gradient products as components
promoting fast learning across multiple axes of generalization.

\vspace{-\baselineskip}
\begin{equation}
    \begin{split}
\nabla_{\Sigma} &= g_{1} + g_{2} +  g_{T} \\ 
    &=  \bar{g}_1 +\bar{g}_2 + \bar{g}_T \\ & -\alpha\left(\bar{H}_{2}\bar{g}_{1}+\bar{H}_{T}\lbrack\bar{g}_1 +\bar{g}_2\rbrack\right) + O\left(\alpha^2 \right)  \label{eq:sigma_gradient_taylor} \\
\end{split}
\end{equation}

The key hyperparameter in \mbox{\sc XG-Reptile} is the number of
inner-loop steps~$K$ representing a trade-off between a manifold
approximation and target step frequency. At small $K$, the 
manifold approximation may be poor, leading to sub-optimal learning. At
large $K$, then improved manifold approximation incurs fewer
target batch steps per epoch, leading to weakened cross-lingual
transfer. In practice, $K$ is set empirically, and Section~\ref{sec:results}
identifies an optimal region for our task.

\mbox{\sc XG-Reptile} can be viewed as generalizing two existing
algorithms. Without the $\mathcal{L}_{T}$ loss, our approach is
equivalent to {Reptile} and lacks cross-lingual alignment. If $K=1$, then \mbox{\sc XG-Reptile} is equivalent to
\mbox{\sc DG-FMAML} \citep{wang-etal-2021-meta-DGMAML} but lacks
intra-task generalization across support batches. Our unification of
these algorithms represent the best of both approaches and
outperforms both techniques within semantic parsing. Another perspective is that \mbox{\sc XG-Reptile} learns
a \textit{regularized optimal manifold}, with immediate cross-lingual capability
as opposed to standard {Reptile} which requires fine-tuning to transfer across tasks.
We identify how this contrast in approaches influences cross-lingual transfer in Section \ref{sec:results}.

%%% Local Variables: 
%%% mode: latex
%%% TeX-master:  "../main"
%%% End:

% !TEX root = ../main.tex

\section{Experimental Design}
\label{sec:exp_design}

We evaluate \mbox{\sc XG-Reptile} against several comparison systems
across multiple languages. Where possible, we re-implement existing
models and use identical data splits to isolate the contribution of
our training algorithm.

\subsection{Data}

We report results on two semantic parsing datasets. First on
\textbf{ATIS} \citep{hemphill-etal-1990-atis}, using the multilingual
version from \citet{sherborne-lapata-2022-zero} pairing utterances in six
languages (English, French, Portuguese, Spanish, German, Chinese) to
SQL queries. ATIS is split into 4,473 training pairs with 493 and 448
examples for validation and testing, respectively. 
We report performance as execution accuracy to test if
predicted SQL queries can retrieve accurate database results.

We also evaluate on \textbf{Spider} \citep{yu-etal-2018-spider}, combining
English and Chinese \citep[CSpider]{Min2019-CSPIDER} versions as a
cross-lingual task. The latter translates all questions to Chinese but
retains the English database. Spider is significantly more
challenging; it contains~10,181 questions and 5,693 unique SQL queries
for 200 multi-table databases over 138 domains. We use the same split
as \citet{wang-etal-2021-meta-DGMAML} to measure generalization to
unseen databases/table-schema during testing. This split uses 8,659
examples from 146 databases for training and 1,034 examples from 20
databases for validation. The test set contains 2,147 examples from 40
held-out databases and is held privately by the authors. To our knowledge, we report the first multilingual approach for Spider by training one model for English and Chinese. Our
challenge is now multi-dimensional, requiring cross-lingual and
cross-domain generalization. Following \citet{yu-etal-2018-spider}, we
report exact set match accuracy for evaluation.

\subsection{Sampling for Generalization}
\label{sec:sampling_for_xgen}

Training for cross-lingual generalization often uses parallel
samples across languages. We illustrate this in
Equation~\eqref{eq:sample1} where~$y_1$ is the equivalent
output for inputs, $x_{1}$, in each language:
\begin{align}
  {\rm EN:}&\left( x_1, y_1 \right)~{\rm DE:}\left( x_1, y_1
        \right)~{\rm ZH:}\left( x_1, y_1 \right) \label{eq:sample1}
\end{align}
However, high sample overlap risks trivializing the task because
models are not learning from new pairs, but instead matching only new
\textit{inputs} to known outputs. A preferable evaluation will test
composition of novel outputs from unseen inputs:
\begin{align}
  {\rm EN:}&\left( x_1, y_1 \right)~{\rm DE:}\left( x_2, y_2 \right)~{\rm ZH:}\left( x_2, y_2 \right) \label{eq:sample2}
\end{align}
Equation~\eqref{eq:sample2} samples exclusive, disjoint datasets for
English and target languages during training. In other words, this
process is \textit{subtractive} e.g.,~a 5\%~sample of German (or Chinese) target
data leaves 95\% of data as the English support. This is similar to
\mbox{K-fold} cross-validation used to evaluate across many data
splits. We sample data for our experiments with
Equation~\eqref{eq:sample2}. It is also possible to use
Equation~\eqref{eq:sample3}, where target samples are also disjoint,
but we find this setup results in too few English examples for
effective
learning.
\begin{align}
  {\rm EN:}&\left( x_1, y_1 \right)~{\rm DE:}\left( x_2, y_2 \right)~{\rm ZH:}\left( x_3, y_3 \right) \label{eq:sample3}
\end{align}

\subsection{Semantic Parsing Models}

We use a Transformer encoder-decoder model
similar to \citet{sherborne-lapata-2022-zero} for our ATIS experiments. We use the same mBART50
encoder \citep{tang2020multilingua-mbart50} and train a Transformer
decoder from scratch to generate SQL.

For Spider, we use the RAT-SQL model \citep{wang-etal-2020-rat} which
has formed the basis of many performant submissions to the
\href{https://yale-lily.github.io/spider}{Spider leaderboard}. RAT-SQL
can successfully reason about unseen databases and table schema using
a novel schema-linking approach within the encoder. We use the version from
\citet{wang-etal-2021-meta-DGMAML} with mBERT
\citep{devlin-etal-2019-BERT} input embeddings for a unified model between
English and Chinese inputs. Notably, RAT-SQL can be over-reliant on lexical similarity features between
input questions and tables \cite{wang-etal-2020-rat}. This raises the
challenge of generalizing to Chinese where such overlap is null.
For fair comparison, we implement identical models as prior work on
each dataset and only evaluate the change in training algorithm. 
This is why we use an mBART50 encoder component for ATIS experiments
and different mBERT input embeddings for Spider experiments.

\subsection{Comparison Systems}

We compare our algorithm against several strong baselines and
adjacent training methods including:
\begin{itemize}
\item[\textbf{Monolingual Training}] A monolingual Transformer
is trained on gold-standard professionally translated data for
  each new language. This is a monolingual upper bound without
  few-shot constraints.\vspace{-0.5em}
\item[\textbf{Multilingual Training}] A multilingual Transformer 
  is trained on the union of all data from the ``Monolingual Training''
  method. This ideal upper bound uses all data in all languages
  without few-shot constraints.\vspace{-0.5em}
\item[\textbf{Translate-Test}] A monolingual Transformer  is trained
  on source English data ($\mathcal{S}_{\rm EN}$). Machine
  translation is used to translate test data from additional languages into
  English. Logical forms are predicted from translated data using the English
  model.\vspace{-0.5em}
\item[\textbf{Translate-Train}] Machine translation is used to translate
  English training data into each target language. A monolingual
  Transformer is trained on translated training data and  logical forms
are predicted   using this model. \vspace{-0.5em}
\item[\textbf{Train-EN$\cup$All}] A Transformer is trained on English
  data and samples from \emph{all} target languages together in a single stage i.e.,~$\mathcal{S}_{\rm EN}\cup\mathcal{S}_{L}$. This is superior to
  training without English (e.g., on $\mathcal{S}_{L}$ only), we
  contrast to this approach for more competitive
  comparison. \vspace{-0.5em}
\item[\textbf{TrainEN$\rightarrow$FT-All}] We first train on English
  support data, $\mathcal{S}_{\rm EN}$, and then fine-tune on target
  samples, $\mathcal{S}_{L}$. \vspace{-0.5em}
\item[\textbf{Reptile-EN$\rightarrow$FT-All}] Initial training uses {Reptile}
  \citep{DBLP:journals/corr/abs-1803-02999-REPTILE} on English support
  data,~$\mathcal{S}_{\rm EN}$, followed by fine-tuning on target
  samples, $\mathcal{S}_{L}$. This is a typical usage of
  {Reptile} for training a low-resource multi-domain parser
  \citep{chen-etal-2020-low-resource-domain-adaptation-task-oriented-parsing}.
\end{itemize}

We also compare to DG-FMAML \citep{wang-etal-2021-meta-DGMAML} as a
special case of \mbox{\sc XG-Reptile} when~\mbox{$K=1$}. Additionally,
we omit pairwise versions of \mbox{\sc XG-Reptile} (e.g., separate models
generalizing from English to individual languages). 
These approaches demand more computation and demonstrated no significant
improvement over a multi-language approach. All Machine Translation uses Google Translate
\citep{gtranslate}.

\subsection{Training Configuration}

\begin{table}[t]
\centering
\small
% \resizebox{\columnwidth}{!}{%
\begin{tabular}{@{}lcc@{}}
\toprule
 & ATIS & Spider \\ \midrule
Batch Size & 10 & 16 \\
Inner Optimizer & \multicolumn{2}{c}{SGD} \\
Inner LR & \multicolumn{2}{c}{$1\times 10^{-4}$}   \\
Outer Optimizer & \multicolumn{2}{c}{Adam { \citep{DBLP:journals/corr/KingmaB14}}} \\
Outer LR & $1\times 10^{-3}$ & $5\times 10^{-4}$ \\
Optimum $K$ & 10 & 3 \\
Max Train Steps & \multicolumn{2}{c}{20,000} \\
Training Time & 12 hours & 2.5 days \\ \bottomrule
\end{tabular}%
% }
\caption{Experimental Hyperparameters for \mbox{\sc XG-Reptile} on ATIS and Spider set primarily by replicating prior work.}
\label{tab:parameter_table}
\vspace{-1em}
\end{table}

Experiments focus on the expansion from English to additional
languages, where we use English as the ``support'' language and
additional languages as ``target''. Key hyperparameters are outlined
in Table \ref{tab:parameter_table}. We train each model using the
given optimizers with early stopping where model selection is through
minimal validation loss for combined support and target
languages. Input utterances are tokenized using SentencePiece
\citep{kudo-richardson-2018-sentencepiece} and Stanza
\citep{qi-etal-2020-stanza} for ATIS and Spider, respectively. All
experiments are implemented in Pytorch on a single V100 GPU. We report
key results for ATIS averaged over three seeds and five random data splits. For Spider, we submit the best singular model from five random splits to the leaderboard.

%%% Local Variables: 
%%% mode: latex
%%% TeX-master:  "../main"
%%% End:

% !TEX root = ../main.tex

\begin{table*}[!ht]
\centering
\resizebox{\textwidth}{!}{%
\footnotesize 
\begin{tabular}{@{}llccccccc@{}}
\toprule
\multicolumn{2}{l}{} & EN & FR & PT & ES & DE & ZH & Target Avg \\ \midrule
\multicolumn{2}{l}{{\sc ZX-Parse}~\citep{sherborne-lapata-2022-zero}} & 76.9 & 70.2 & 63.4 & 59.7 & 69.3 & 60.2 & 64.6$~\pm~$5.0 \\
\multicolumn{2}{l}{Monolingual Training} & 77.2 & 67.8 & 66.1 & 64.1 & 66.6 & 64.9 & 65.9$~\pm~$1.4 \\
\multicolumn{2}{l}{Multilingual Training} & 73.9 & 72.5 & 73.1 & 70.4 & 72.0 & 70.5 & 71.7$~\pm~$1.2 \\
\multicolumn{2}{l}{Translate-Train} &  --- & 55.9 & 56.1 & 57.1 & 60.1 & 56.1 & 57.1$~\pm~$1.8 \\
\multicolumn{2}{l}{Translate-Test} &  --- & 58.2 & 57.3 & 57.9 & 56.9 & 51.4 & 56.3$~\pm~$2.8 \\ \midrule
\multirow{4}{*}{@1\%} & Train-EN$\cup$All & 69.7$~\pm~$1.4 & 44.0$~\pm~$3.5 & 42.2$~\pm~$3.7 & 38.3$~\pm~$6.8 & 45.8$~\pm~$2.6 & 41.7$~\pm~$3.6 & 42.4$~\pm~$2.8 \\
 & Train-EN$\rightarrow$FT-All & 71.2$~\pm~$2.3 & 53.3$~\pm~$5.2 & 49.7$~\pm~$5.4 & 56.1$~\pm~$2.7 & 52.5$~\pm~$6.7 & 39.0$~\pm~$4.0 & 50.1$~\pm~$6.6 \\
 & Reptile-EN$\rightarrow$FT-All & 73.2$~\pm~$0.7 & 58.9$~\pm~$4.8 & 54.8$~\pm~$3.4 & 52.8$~\pm~$4.4 & 60.6$~\pm~$3.6 & 41.7$~\pm~$4.0 & 53.8$~\pm~$7.4 \\
 & \mbox{\sc XG-Reptile} & {\bf73.8}$~\pm~$0.3 & {\bf70.4}$~\pm~$1.8 & {\bf70.8}$~\pm~$0.7 & {\bf68.9}$~\pm~$2.3 & {\bf69.1}$~\pm~$1.2 & {\bf68.1}$~\pm~$1.2 & {\bf69.5}$~\pm~$1.1 \\ \midrule
\multirow{4}{*}{@5\%} & Train-EN$\cup$All & 67.3$~\pm~$1.6 & 55.2$~\pm~$4.5 & 54.7$~\pm~$4.5 & 44.4$~\pm~$4.5 & 55.8$~\pm~$2.9 & 52.3$~\pm~$4.3 & 52.5$~\pm~$4.7 \\
 & Train-EN$\rightarrow$FT-All & 69.2$~\pm~$1.9 & 58.9$~\pm~$5.3 & 54.8$~\pm~$5.4 & 52.8$~\pm~$4.5 & 60.6$~\pm~$6.5 & 41.7$~\pm~$9.5 & 53.8$~\pm~$7.4 \\
 & Reptile-EN$\rightarrow$FT-All & 69.5$~\pm~$1.8 & 65.3$~\pm~$3.8 & 61.3$~\pm~$6.0 & 59.6$~\pm~$2.6 & 64.9$~\pm~$5.1 & 56.9$~\pm~$9.2 & 61.6$~\pm~$3.6 \\
 & \mbox{\sc XG-Reptile} & {\bf74.4}$~\pm~$1.3 & {\bf73.0}$~\pm~$0.9 & {\bf71.6}$~\pm~$1.1 & {\bf71.6}$~\pm~$0.7 & {\bf71.1}$~\pm~$0.6 & {\bf69.5}$~\pm~$0.5 & {\bf71.4}$~\pm~$1.3 \\ \midrule
\multirow{4}{*}{@10\%} & Train-EN$\cup$All & 65.7$~\pm~$1.9 & 61.5$~\pm~$1.7 & 62.1$~\pm~$2.3 & 53.7$~\pm~$3.2 & 62.7$~\pm~$2.3 & 60.6$~\pm~$2.4 & 60.1$~\pm~$3.7 \\
 & Train-EN$\rightarrow$FT-All & 67.4$~\pm~$1.9 & 63.8$~\pm~$5.8 & 60.3$~\pm~$5.3 & 59.6$~\pm~$4.0 & 64.5$~\pm~$6.5 & 58.4$~\pm~$6.4 & 61.3$~\pm~$2.7 \\
 & Reptile-EN$\rightarrow$FT-All & 72.8$~\pm~$1.8 & 66.3$~\pm~$4.2 & 64.6$~\pm~$4.9 & 62.3$~\pm~$6.4 & 66.6$~\pm~$5.0 & 60.7$~\pm~$3.6 & 64.1$~\pm~$2.6 \\
 & \mbox{\sc XG-Reptile} & {\bf75.8}$~\pm~$1.3 & {\bf74.2}$~\pm~$0.2 & {\bf72.8}$~\pm~$0.6 & {\bf72.1}$~\pm~$0.7 & {\bf73.0}$~\pm~$0.6 & {\bf72.8}$~\pm~$0.5 & {\bf73.0}$~\pm~$0.8 \\ \bottomrule
\end{tabular}
}
\caption{Denotation accuracy using varying learning algorithms including \mbox{\sc XG-Reptile} at 1\%, 5\%, and 10\% sampling rates for target dataset size relative to support dataset for ATIS. We report for \emph{English},~\emph{French},~\emph{Portuguese},~\emph{Spanish},~
\emph{German} and \emph{Chinese}. \emph{Target Avg} reports the average denotation accuracy across non-English languages $\pm$ standard deviation across languages. For few-shot experiments, we also report the standard deviation ($\pm$) across random samples.  Best few-shot results per language are bolded.}
\label{tab:xg_reptile_atis_1}
\end{table*}

\section{Results and Analysis}
\label{sec:results}

\begin{figure*}[ht]
    \centering
    \includegraphics[width=\textwidth]{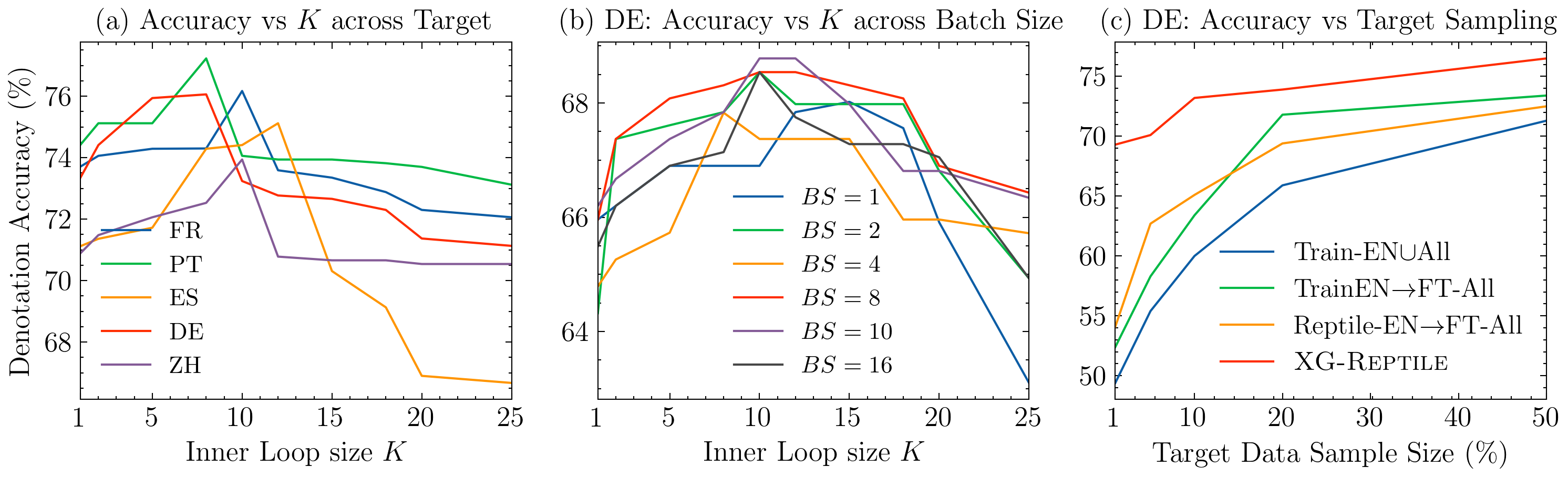}
    \caption{Ablation Experiments on ATIS (a) accuracy against inner loop size $K$ across languages, (b) accuracy against $K$ for German when varying batch size, and (c) accuracy against dataset sample size relative to support dataset from 1\% to 50\% for German. For~(b), the $K=1$ case is equivalent to DG-FMAML \citep{wang-etal-2021-meta-DGMAML}.}
    \label{fig:k_exp_1_2_3}
    \vspace{-1em}
\end{figure*}

We contrast \mbox{\sc XG-Reptile} to baselines for ATIS in
Table~\ref{tab:xg_reptile_atis_1} and present further analysis within
Figure~\ref{fig:k_exp_1_2_3}. Results for the multi-domain Spider are
shown in Table~\ref{tab:xgr_spider_1}. Our findings
support our hypothesis that \mbox{\sc XG-Reptile} is a superior
algorithm for jointly training a semantic parser and encouraging
cross-lingual generalization with improved sample efficiency. Given the same
data, \mbox{\sc XG-Reptile} produces more mutually beneficial
parameters for both model requirements with only modifications to the
training loop.

\vspace{-\baselineskip}
\paragraph{Comparison across Generalization Strategies}

We compare \mbox{\sc XG-Reptile} to established learning algorithms in
Table \ref{tab:xg_reptile_atis_1}. Across baselines, we find that
single-stage training, i.e., \emph{Train-EN$\cup$All} or
machine-translation based models, perform below two-stage
approaches. The strongest competitor is the
\emph{Reptile-EN$\rightarrow$FT-All} model, highlighting the
effectiveness of Reptile for single-task generalization
\citep{kedia-etal-2021-beyond-reptile}. However, \mbox{\sc XG-Reptile}
performs above all baselines across sample rates. Practically, 1\%,
5\%, 10\% correspond to 45, 225, and 450 example pairs, respectively. We
identify significant improvements ($p<0.01$; relative to the closest model using an independent t-test) in cross-lingual transfer through jointly learning to parse and multi-language generalization while maintaining single-stage training efficiency.

Compared to the upper bounds, \mbox{\sc XG-Reptile} performs above
\emph{Monolingual Training} at $\ge 1\%$ sampling, which further
supports the prior benefit of multilingual modeling
\citep{arch-for-neural-multisp-Susanto2017}.
\emph{Multilingual Training} is only marginally stronger than \mbox{\sc
  XG-Reptile} at 1\% and 5\% sampling despite requiring many more examples. \mbox{\sc
  XG-Reptile}@10\% improves on this model by an average~$+$1.3\%. Considering that our upper bound uses $10\times$ the data of
\mbox{\sc XG-Reptile}@10\%, this accuracy gain highlights the benefit
of explicit cross-lingual generalization. This is
consistent at higher sample sizes (see Figure
\ref{fig:k_exp_1_2_3}(c) for German).

At the smallest sample size, \mbox{\sc XG-Reptile}@1\%, demonstrates a
$+$12.4\% and $+$13.2\% improvement relative to \emph{Translate-Train} and
\emph{Translate-Test}. Machine translation is often viable for cross-lingual transfer \citep{conneau2018xnli}. However, we find that mistranslation of named entities
incurs an exaggerated parsing penalty -- leading to inaccurate logical forms
\citep{sherborne-etal-2020-bootstrapping}. This suggests that
sample quality has an exaggerated influence on semantic parsing
performance. When training \mbox{\sc XG-Reptile} with MT
data, we also observe a lower Target-language average of 66.9\%. 
This contrast further supports
the importance of sample quality in our context.

 \mbox{\sc XG-Reptile} improves cross-lingual generalization across all languages at
 equivalent and lower sample sizes. At 1\%, it
 improves by an average $+$15.7\% over the closest model,
 \emph{Reptile-EN$\rightarrow$FT-All}. Similarly, at 5\%, we find
 $+$9.8\% gain, and at 10\%, we find $+$8.9\% relative to the closest
 competitor. Contrasting across sample sizes --- our best approach is
 @10\%, however, this is $+$3.5\%~above @1\%, suggesting that smaller
 samples could be sufficient if 10\% sampling is unattainable. This
 relative stability is an improvement compared to the 17.7\%, 11.2\% or
 10.3\% difference between @1\% and @10\% for other models. This
 implies that \mbox{\sc XG-Reptile} better utilizes smaller samples
 than adjacent methods.

Across languages at 1\%, \mbox{\sc XG-Reptile} improves primarily for
languages dissimilar to English \citep{ahmad-etal-2019-difficulties}
to better minimize the cross-lingual transfer gap. For Chinese
(ZH), we see that \mbox{\sc XG-Reptile}@1\% is $+$26.4\% above the
closest baseline. This contrasts with the smallest gain, $+$8.5\%~for
German, with greater similarity to English. Our improvement also
yields less variability across target languages --- the
standard deviation across languages for \mbox{\sc XG-Reptile}@1\% is
1.1, compared to 2.8 for \emph{Train-EN$\cup$All} or 7.4 for 
\emph{Reptile-EN$\rightarrow$FT-All}.

We can also compare to {\sc ZX-Parse}, the method of
\citet{sherborne-lapata-2022-zero} which engineers cross-lingual latent
alignment for zero-shot semantic parsing without data in target
languages. With 45 samples per target language, \mbox{\sc
  XG-Reptile}@1\% improves by an average of $+$4.9\%.  \mbox{\sc
  XG-Reptile} is more beneficial for distant languages --
cross-lingual transfer penalty between English and Chinese is $-$12.3\%
for {\sc ZX-Parse} compared to $-$5.7\% in our case. While these
systems are not truly comparable, given different data requirements,
this contrast is practically useful for comparison between zero- and
few-shot localization.

\paragraph{Influence of~$K$ on Performance}
In Figure~\ref{fig:k_exp_1_2_3}(a) we study how variation in the key
hyperparameter $K$, the size of the inner-loop in
Algorithm~\ref{xgr-algo} or the number of batches used to approximate
the \emph{optimal task manifold} influences model performance across
languages (single run at 5\% sampling).  When \mbox{$K=1$}, the model learns
generalization from batch-wise similarity which is equivalent to
\emph{DG-FMAML} \citep{wang-etal-2021-meta-DGMAML}. We empirically
find that increasing~$K$ beyond one benefits performance by
encouraging cross-lingual generalization with the \emph{task} over a
single \emph{batch}, and it is, therefore, beneficial to align an
out-of-domain example with the overall \textit{direction} of training.
However, as theorized in Section~\ref{sec:methodology}, increasing $K$
also decreases the frequency of the outer step within an epoch leading
to poor cross-lingual transfer at high~$K$. This trade-off yields an
optimal operating regime for this hyper-parameter. We use $K=10$ in
our experiments as the center of this region. Given this setting
of~$K$, the target sample size must be 10\% of the support sample size
for training in a single epoch. However,
Table~\ref{tab:xg_reptile_atis_1} identifies \mbox{\sc XG-Reptile} as
the most capable algorithm for ``over-sampling'' smaller target
samples for resource-constrained generalization.

\paragraph{Influence of Batch size on performance}
We consider two further case studies to analyze \mbox{\sc
  XG-Reptile} performance. For clarity, we focus on German; however,
these trends are consistent across all target languages.
Figure~\ref{fig:k_exp_1_2_3}(b) examines if the effects of
cross-lingual transfer within \mbox{\sc XG-Reptile} are sensitive to
batch size during training  (single run at 5\% sampling). A dependence between $K$ and batch size
could imply that the desired inter-task and cross-lingual
generalization outlined in Equation~\eqref{eq:sigma_gradient_taylor}
is an unrealistic, edge-case phenomenon. This is not the
case, and a trend of optimal $K$ setting is consistent across many
batch sizes. This suggests that $K$ is an independent hyper-parameter
requiring tuning alongside existing experimental settings.

\paragraph{Performance across Larger Sample Sizes} We consider a wider
range of target data sample sizes between 1\% to 50\% in
Figure~\ref{fig:k_exp_1_2_3}(c). We observe that baseline approaches
converge to between 69.3\% and 73.9\% at 50\% target sample
size. Surprisingly, the improvement of \mbox{\sc XG-Reptile} is
retained at higher sample sizes with an accuracy of 76.5\%. The benefit of \mbox{\sc
  XG-Reptile} is still greatest at low sample sizes with $+$5.4\%
improvement at 1\%; however, we maintain a $+$2.6\% gain over the
closest system at 50\%. While low sampling is the most
economical, the consistent benefit of \mbox{\sc XG-Reptile}
suggests a promising strategy for other cross-lingual tasks.

\begin{table}[t]
\small
\centering
\begin{tabular}{@{}llcc|cc@{}}
\toprule
\multicolumn{2}{l}{} & \multicolumn{2}{c}{EN} & \multicolumn{2}{c}{ZH} \\ \midrule
\multicolumn{2}{l}{} & Dev & Test & Dev & Test \\ \midrule
\multicolumn{6}{l}{\textit{Monolingual}} \\ \midrule
\multicolumn{2}{l}{\makecell[l]{DG-MAML }} & \textbf{68.9} &  \textbf{65.2} & \textbf{50.4} &  \textbf{46.9} \\
\multicolumn{2}{l}{\makecell[l]{DG-FMAML }} & 56.8 &  --- & 32.5 &  --- \\
\multicolumn{2}{l}{\mbox{\sc XG-Reptile}} & 63.5 &  --- & 48.9 &  --- \\ \midrule
\multicolumn{6}{l}{\textit{Multilingual}} \\ \midrule
\mbox{\sc XG-Reptile} 
 & @1\% & 56.8 &  56.5 & 47.0 &  45.6 \\
 & @5\% & \textbf{59.6} &  58.1 & 47.3 &  45.6 \\
 & @10\% & 59.2 &  \textbf{59.7} & \textbf{48.0} &  \textbf{46.0} \\
 \bottomrule
\end{tabular}
\caption{Exact set match accuracy for RAT-SQL trained on Spider (English) and CSpider (Chinese) comparing \mbox{\sc XG-Reptile} to DG-MAML and DG-FMAML \citep{wang-etal-2021-meta-DGMAML}. We experiment with sampling between 1\% to 10\% of Chinese examples relative to English. Monolingual and multilingual best results are bolded.
}
\label{tab:xgr_spider_1}
\vspace{-1.5em}
\end{table}

\paragraph{Learning Spider and CSpider}

Our results on Spider and CSpider are shown in
Table~\ref{tab:xgr_spider_1}. We compare \mbox{\sc XG-Reptile}
to monolingual approaches from  \citet{wang-etal-2021-meta-DGMAML}
and discuss cross-lingual results when sampling between 1\% to 10\% of CSpider target
during training.

In the \emph{monolingual setting}, \mbox{\sc XG-Reptile} shows
significant improvement ($p<0.01$; using an independent samples t-test) compared to DG-FMAML with $+$6.7\% for English and $+$16.4\% for Chinese dev accuracy. This further supports
our claim that generalizing with a \textit{task manifold} is superior
to batch-level generalization. 

Our results are closer to DG-MAML \citep{wang-etal-2021-meta-DGMAML}, a higher-order meta-learning method requiring computational resources and training times exceeding $4\times$ the requirements for {\sc XG-Reptile}. \mbox{\sc XG-Reptile} yields accuracies $-5.4\%$ and $-1.5\%$ below DG-MAML for English and Chinese, where DG-FMAML performs much lower at $-12.1\%$ (EN) and $-17.9\%$ (ZH). 
Our results suggest that \mbox{\sc XG-Reptile} is a superior first-order meta-learning
algorithm rivaling prior work with greater computational demands.\footnote{We compare against DG-MAML as the
best \emph{public} available model on the \href{https://taolusi.github.io/CSpider-explorer/}{CSpider leaderboard}
at the time of writing.}

In the multilingual setting, we observe that
\mbox{\sc XG-Reptile} performs competitively using as little as 1\% of
Chinese examples. While training sampling 1\% and 5\% perform similarly -- the best model sees 10\% of CSpider samples during
training to yield accuracy only $-$0.9\%~(test) below the monolingual DG-MAML model.
While performance does not match monolingual models, the multilingual
approach has additional utility in serving more users. As a zero-shot
setup, predicting SQL from CSpider inputs through the model trained
for English, yields 7.9\%~validation accuracy, underscoring that
cross-lingual transfer for this dataset is \mbox{non-trivial}.

Varying the target sample size demonstrates more variable effects for
Spider compared to ATIS. Notably, increasing the sample size yields
poorer English performance beyond the optimal \mbox{\sc
  XG-Reptile}@5\% setting for English. This may be a consequence of
the cross-database challenge in Spider --- information shared across
languages may be less beneficial than for single-domain ATIS. The
least performant model for both languages is \mbox{\sc
  XG-Reptile}@1\%. Low performance here for Chinese can be
expected, but the performance for English is surprising. We suggest that
this result is a consequence of ``over-sampling'' of the target
data disrupting the overall training process. That is, for 1\%
sampling and optimal $K=4$, the target data is ``over-used''
$25\times$ for each epoch of support data. We further observe
diminishing benefits for English with additional Chinese
samples. While we trained a competitive parser with minimal
Chinese data, this effect could be a consequence of how
RAT-SQL cannot exploit certain English-oriented learning features (e.g., lexical similarity scores). Future
work could explore cross-lingual strategies to unify entity modeling for
improved feature sharing.

\paragraph{Visualizing the Manifold}

\begin{figure*}[ht]
    \centering
    \includegraphics[width=\textwidth]{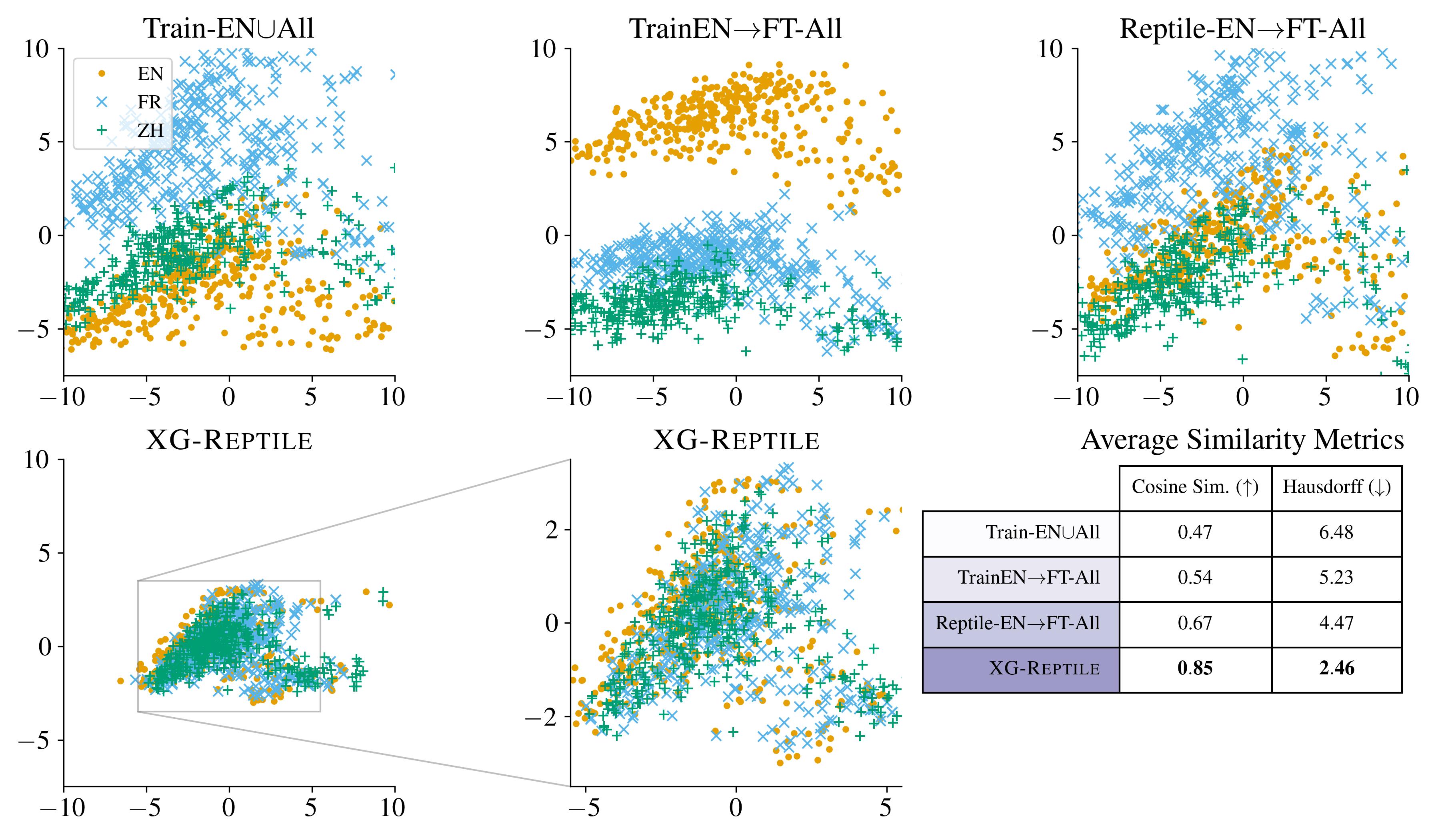}
    \caption{PCA Visualizations of sentence-averaged encodings for English (EN), French (FR) and Chinese (ZH) from the ATIS test set (@1\% sampling from Table \ref{tab:xg_reptile_atis_1}). We identify the regularized weight manifold which improves cross-lingual transfer using \mbox{\sc XG-Reptile}. We also improve in two similarity metrics averaged across languages.}
    \label{fig:pca}
    \vspace{-1em}
\end{figure*}
 
Analysis of \mbox{\sc XG-Reptile} in Section \ref{sec:methodology} relies on a theoretical basis that first-order meta-learning creates a dense high-likelihood sub-region in the parameters (i.e. \textit{optimal manifold}). 
Under these conditions, representations of new domains should cluster within the manifold to allow for rapid adaptation with minimal samples. This contrasts with methods without meta-learning, which provide no guarantees of representation density. However, metrics in Table \ref{tab:xg_reptile_atis_1} and \ref{tab:xgr_spider_1} do not directly explain if this expected effect arises. To this end, we visualize ATIS test set encoder outputs using PCA \citep{doi:10.1137/090771806} in Figure \ref{fig:pca}. We contrast English (support) and French and Chinese as the most and least similar target languages. Using PCA allows for direct interpretation of low-dimensional distances across approaches. Cross-lingual similarity is a proxy for manifold alignment -- as our goal is accurate cross-lingual transfer from closely aligned representations from source and target languages \citep{xia-etal-2021-metaxl, sherborne-lapata-2022-zero}.

Analyzing Figure \ref{fig:pca}, we observe meta-learning methods (\emph{Reptile-EN$\rightarrow$FT-All},~\mbox{\sc XG-Reptile}) to fit target languages closer to the support (English, yellow circle). In contrast, methods not utilizing meta-learning (\emph{Train-EN$\cup$All},~\emph{Train-EN$\rightarrow$FT-All}) appear less ordered and with weaker representation overlap. Encodings from \mbox{\sc XG-Reptile} are less separable across languages and densely clustered, suggesting the regularized manifold hypothesized in Section \ref{sec:methodology} ultimately yields improved cross-lingual transfer. Visualizing encodings from English in the \emph{Reptile-EN} model \textit{before} fine-tuning produces a similar cluster (not shown), however, required fine-tuning results in ``spreading'' leading to less cross-lingual similarity. 

We also quantitatively examine the average encoding change in Figure \ref{fig:pca} using Cosine similarity and Hausdorff distance \citep{patra-etal-2019-bilingual} between English and each target language. Cosine similarity is measured pair-wise across parallel inputs in each language to gauge similarity from representations with equivalent SQL outputs. As a measure of mutual proximity between sets, Hausdorff distance denotes a worst-case distance between languages to measure more general ``closeness''. Under both metrics, \mbox{\sc XG-Reptile} yields the best performance with the most substantial pair-wise similarity and Hausdorff similarity. These indicators for cross-lingual similarity further support the observation that our expected behavior is legitimately occurring during training. 

Our findings better explain \textit{why} our \mbox{\sc XG-Reptile} performs above other training algorithms. Specifically, our results suggest that \mbox{\sc XG-Reptile} learns a \textit{regularized manifold} which produces stronger cross-lingual similarity and improved parsing compared to Reptile \textit{fine-tuning a manifold}. This contrast will inform future work for cross-lingual meta-learning where \mbox{\sc XG-Reptile} can be applied. 

\paragraph{Error Analysis}

\begin{figure}[t]
\centering
\begin{tabulary}{\columnwidth}{@{}LL@{}}
\toprule
EN & {\footnotesize Show me all flights from San Jose to Phoenix} \\ \midrule
FR & {\footnotesize Me montrer tous les vols de San Jos\`e \'a Phoenix } \\ \midrule
$\times$ & {\footnotesize SELECT DISTINCT flight\_1.flight\_id FROM flight flight\_1, airport\_service airport\_service\_1, city city\_1, airport\_service airport\_service\_2, city city\_2 WHERE flight\_1.from\_airport = airport\_service\_1.airport\_code AND airport\_service\_1.city\_code = city\_1.city\_code AND city\_1.city\_name = \colorbox{red}{'SAN FRANCISCO'} AND flight\_1.to\_airport = airport\_service\_2.airport\_code AND airport\_service\_2.city\_code = city\_2.city\_code AND city\_2.city\_name =  \colorbox{red}{'PHILADELPHIA'};} \\ \midrule
\checked & {\footnotesize SELECT DISTINCT flight\_1.flight\_id FROM flight flight\_1, airport\_service airport\_service\_1, city city\_1, airport\_service airport\_service\_2, city city\_2 WHERE flight\_1.from\_airport = airport\_service\_1.airport\_code AND airport\_service\_1.city\_code = city\_1.city\_code AND city\_1.city\_name =  \colorbox{ForestGreen}{'SAN JOSE'} AND flight\_1.to\_airport = airport\_service\_2.airport\_code AND airport\_service\_2.city\_code = city\_2.city\_code AND city\_2.city\_name =  \colorbox{ForestGreen}{'PHOENIX'};} \\ \bottomrule
\end{tabulary}
\caption{Contrast between SQL from a French input from ATIS for  \emph{Train-EN$\cup$All} and \mbox{\sc XG-Reptile}. The entities ``San Jos\'{e}'' and ``Phoenix'' are not observed in the 1\% sample of French data but are mentioned in the English support data. The  \emph{Train-EN$\cup$All} approach fails to connect attributes seen in English when generating SQL from French inputs ($\times$). Training with \mbox{\sc XG-Reptile} better leverages support data to generate accurate SQL from other languages (\checked).
}
\label{tab:entity-error}
\vspace{-1em}
\end{figure}

We can also examine \textit{where} the improved cross-lingual transfer influences parsing performance. Similar to Figure \ref{fig:pca}, we consider the results of models using 1\% sampling as the worst-case performance and examine where \mbox{\sc XG-Reptile} improves on other methods on the test set (448 examples) over five languages.

Accurate semantic parsing requires sophisticated entity handling to translate mentioned proper nouns from utterance to logical form. In our few-shot sampling scenario, \textit{most} entities will appear in the English support data (e.g. ``Denver'' or ``American Airlines''), and \textit{some} will be mentioned within the target language sample (e.g. ``Mine\'apolis'' or ``Nueva York'' in Spanish). These samples cannot include all possible entities -- effective cross-lingual learning must ``connect'' these entities from the support data to the target language -- such that these names can be parsed when predicting SQL from the target language. As shown in Figure \ref{tab:entity-error}, the failure to recognize entities from support data, for inference on target languages, is a critical failing of all models besides \mbox{\sc XG-Reptile}. 

The improvement in cross-lingual similarity using \mbox{\sc XG-Reptile} expresses a specific improvement in entity recognition. Compared to the worst performing model, \emph{Train-EN$\cup$All}, 55\% of improvement accounts for handling entities absent from the 1\% target sample but present in the 99\% English support data. While \mbox{\sc XG-Reptile} can generate accurate SQL, other models are limited in expressivity to fall back on using seen entities from the 1\% sample. This notably accounts for 60\% of improvement in parsing Chinese, with minimal orthographic overlap to English, indicating that \mbox{\sc XG-Reptile} better leverages support data without reliance on token similarity. In 48\% of improved parses, entity mishandling is the \textit{sole error} -- highlighting how limiting poor cross-lingual transfer is for our task. 

Our model also improves handling of novel \textit{modifiers} (e.g. ``on a weekday'', ``round-trip'') absent from target language samples. Modifiers are often realized as additional sub-queries and filtering logic in SQL outputs. Comparing \mbox{\sc XG-Reptile} to  \emph{Train-EN$\cup$All}, 33\% of improvement is related to modifier handling. Less capable systems fall back on modifiers observed from the target sample or ignore them entirely to generate inaccurate SQL. 

While \mbox{\sc XG-Reptile} better links parsing knowledge from English to target languages -- the problem is not solved. Outstanding errors in all languages primarily relate to query complexity, and the cross-lingual transfer gap is not closed. Furthermore, our error analysis suggests a future direction for optimal sample selection to minimize the error from interpreting unseen phenomena.

%%% Local Variables: 
%%% mode: latex
%%% TeX-master:  "../main"
%%% End:

\section{Conclusion}
\label{sec:conclusion}

We propose \mbox{\sc XG-Reptile}, a meta-learning
algorithm for few-shot cross-lingual generalization in semantic
parsing. \mbox{\sc XG-Reptile} is able to better utilize fewer samples
to learn an economical multilingual semantic parser with minimal cost
and improved sample efficiency. Compared to adjacent training algorithms
and zero-shot approaches, we obtain more accurate and
consistent logical forms across languages similar and dissimilar to
English. Results on ATIS show clear benefit across many languages and
results on Spider demonstrate that \mbox{\sc XG-Reptile} is effective
in a challenging cross-lingual and cross-database scenario. We focus
our study on semantic parsing, however, this algorithm could be
beneficial in other low-resource cross-lingual tasks. In future work
we plan to examine how to better align entities in low-resource
languages to further improve parsing accuracy.

%%% Local Variables: 
%%% mode: latex
%%% TeX-master:  "../main"
%%% End:

\section*{Acknowledgements}
We thank the action editor and anonymous reviewers for their
constructive feedback. The authors also thank Nikita Moghe, Seraphina
Goldfarb-Tarrant, Ondrej Bohdal and Heather Lent for their insightful
comments on earlier versions of this paper. We gratefully acknowledge
the support of the UK Engineering and Physical Sciences Research
Council (grants EP/L016427/1 (Sherborne) and EP/W002876/1 (Lapata))
and the European Research Council (award 681760, Lapata).

\bibliography{tacl2021}
\bibliographystyle{acl_natbib}

\end{document}